%% file: paper.tex
\pdfoutput=1

\documentclass[11pt]{article}

\usepackage{acl}

\usepackage{times}
\usepackage{latexsym}

\usepackage[utf8]{inputenc}

\usepackage{microtype}

\usepackage{graphicx}

\usepackage{amssymb}
\usepackage{amsmath}
\usepackage{enumitem}
\usepackage{siunitx}
\usepackage{svg}
\usepackage{tabularx}
\usepackage[font=footnotesize]{caption}

\usepackage{physics}
\usepackage{tikz}
\usepackage{mathdots}
\usepackage{yhmath}
\usepackage{cancel}
\usepackage{color}
\usepackage{array}
\usepackage{multirow}
\usepackage{amssymb}
\usepackage{gensymb}
\usepackage{booktabs}

\usepackage{colortbl}
\usepackage{caption}
\usepackage{subcaption}
\usepackage{vcell}

\usetikzlibrary{fadings}
\usetikzlibrary{patterns}
\usetikzlibrary{shadows.blur}
\usetikzlibrary{shapes}

\title{Transformer with Tree-order Encoding for Neural Program Generation}

\author{
  Klaudia-Doris Thellmann\footnotemark \\
  TU Dresden\\\And
  Bernhard Stadler\\
  TU Dresden\\\AND
  Ricardo Usbeck\\
  Fraunhofer IAIS\\\And
  Jens Lehmann \\
  Fraunhofer IAIS\\
}

\begin{document}
\maketitle
\begin{abstract}
While a considerable amount of semantic parsing approaches have employed RNN architectures for code generation tasks, there have been only few attempts to investigate the applicability of Transformers for this task.
Including hierarchical information of the underlying programming language syntax has proven to be effective for code generation.
Since the positional encoding of the Transformer can only represent positions in a flat sequence, we have extended the encoding scheme to allow the attention mechanism to also attend over hierarchical positions in the input.
Furthermore, we have realized a decoder based on a restrictive grammar graph model to improve the generation accuracy and ensure the well-formedness of the generated code.
While we did not surpass the state of the art, our findings suggest that employing a tree-based positional encoding in combination with a shared natural-language subword vocabulary improves generation performance over sequential positional encodings.\footnote{This paper was authored in late 2020 and early 2021 for the most part.}
\end{abstract}

\section{Introduction}
\label{sec:introduction}
Automatically generating source code from instructions in natural language can reduce the human-machine language barrier.
Efforts on overcoming this barrier have led to numerous semantic parsing approaches ranging from statistical semantic parsing with focus on inducing rules~\cite{zelle_learning_1996} or probabilistic grammars~\cite{zettlemoyer_learning_2005} to neural semantic parsing approaches based on encoder-decoder architectures, which have proven to be effective in mapping natural language into a formal meaning representation such as logical forms~\cite{jia_liang_2016,suhr_learning_2018} or general-purpose programming languages (GPPL)~\cite{ling_2016,iyer_mapping_2018,yin_2017,yin_tranx_2018,iyer_learning_2019}. 

Recent syntax-driven neural semantic parsing approaches integrated compositional structures of logical forms~\cite{dong_2016} or, in the case of GPPL, the underlying syntax~\cite{rabinovich_2017,yin_2017,hayati_2018,sun_2019,xu_2020}, which substantially improved generation accuracy.
However, existing grammar-constrained decoding approaches for a GPPL like Python either use a manually defined list of production rules~\cite{yin_2017} or the ASDL~\cite{wang_zephyr_1997} abstract grammar definition~\cite{psf_2018} used by the CPython implementation of the Python language~\cite{rabinovich_2017,yin_tranx_2018}.
The manual approach allows for flexibility in modeling, but requires manually checking and possibly re-modeling the rules with every new release, which is prone to human error.
Although the ASDL eliminates these drawbacks, it is not sufficiently restrictive and therefore cannot guarantee the generation of well-formed, i.e., parsable code.
In compilers and interpreters, this is not problematic because some component in the pipeline consisting of parser, static checker and dynamic checker catches each of these types of error~\cite{aho_compilers_2007}.
In neural semantic parsing, however, it unnecessarily allows the generation of invalid code.

Building on this finding, we propose a syntax-driven neural semantic parsing approach employing a more restrictive grammar model for the task of generating well-formed code from instructions or descriptions in natural language as depicted in Table~\ref{tab:ds-examples}.
\begin{table*}[htbp]
{
\centering
\input{datasets}
}
\caption{Examples of natural language specifications or questions and their corresponding formal meaning representation from the datasets used in this work: Atis~\cite{dahl_1994}, Geo~\cite{tang_mooney_2001}, Hearthstone~\cite{ling_2016} and CoNaLa~\cite{conala_yin_2018}}
\label{tab:ds-examples}
\end{table*}
Most syntax-driven neural semantic parsing approaches employ encoder-decoder architectures based on RNNs~\cite[i.a.]{dong_2016,rabinovich_2017,hayati_2018,iyer_learning_2019}.
With the Transformer~\cite{vaswani_2017}, a more powerful generation of neural architectures was introduced, surpassing RNNs in a variety of NLP tasks, including machine translation~\cite{nguyen_2020,chen_2018} and constituency parsing~\cite{wang_2019,harer_2019}.
Unlike RNNs, Transformers rely solely on self-attention and therefore can capture long-range context dependencies while being easy to parallellize.

Since the Transformer is invariant to sequence ordering, it is of particular importance to explicitly include position information of sequence tokens through learned or fixed positional encoding schemes~\cite{vaswani_2017,shiv_2019} or biased attention weights~\cite{shaw_2018,nguyen_2020}.
While the inclusion of hierarchical information has proven to be effective for code generation, the positional encoding by \cite{vaswani_2017} can only represent positions in a flat sequence.
Taking up this thought, we propose a position encoding scheme for the Transformer which allows the attention mechanism to also attend over hierarchical positions in the input.
Similar to~\cite{shiv_2019}, we also rely on paths in an abstract tree representation of input code snippets to encode hierarchical code structure. 
However,~\cite{shiv_2019}'s encoding, cannot be applied to trees of variable width since the depth and width of the encoding is limited. 

Our contributions are the following:
\begin{enumerate}[label=\arabic*)]
    \item A restrictive grammar graph model, which can be automatically generated from any tree-like data, not requiring an ASDL with the only condition that ordered typed trees are available as input.
    The grammar model is used to determine the hierarchical relations for the tree encoding, but can also be used for constrained decoding, ensuring the generation of parsable code. 
    \item A tree encoding scheme based on the sinusoidal encoding introduced by~\cite{vaswani_2017} to enable the Transformer model to learn hierarchical relations in an ordered tree of arbitrary width.
\end{enumerate}
We evaluated our approach on several benchmark datasets for code generation including Hearthstone~\cite{ling_2016}, CoNaLa~\cite{conala_yin_2018}, Atis~\cite{dahl_1994} and Geo~\cite{tang_mooney_2001}. Experimental results show that he tree encoding performs better than the sequential encoding used by the original Transformer architecture on the Hearthstone dataset.
On the CoNaLa dataset, we achieve improvements when employing a separate subword vocabulary for the string literals extracted from the code snippets.
Our code is publicly available at \url{https://github.com/SmartDataAnalytics/codeCAI}.

\section{Related Work}

A large variety of syntax-driven neural semantic parsing approaches generate abstract syntax trees (ASTs) by predicting a sequence of grammar rules.
Yin et al. \cite{yin_2017,yin_tranx_2018}, Rabinovich et al. \cite{rabinovich_2017} and Sun et al. \cite{sun_2019} define a grammar model that captures the syntax of the target programming language and generate ASTs based on a series of predicted actions, i.e., apply a production rule to expand a non-terminal node or populate a terminal node. 
Most of these approaches derive the underlying grammar model from the official Python abstract grammar definition (ASDL).
However, abstract grammar definitions are designed for compilation or interpretation~\cite{aho_compilers_2007} and not for code generation and therefore, as in the case of the Python grammar, allow syntax trees that do not correspond to parsable code.
We counteract this drawback by creating a more restrictive grammar model from the parsable code samples of the training dataset.
Based on this grammar model, we generate ASTs by predicting the next AST node on the input tree path in depth-first preorder.

To capture the compositional structure of logical forms,~\cite{dong_2016} propose a tree decoder that recursively generates sub-trees, conveying hierarchical information by feeding the parent node's hidden representation as an additional input.
Expanding on this, \cite{rabinovich_2017} realize a grammar-based modular decoder with a dynamically determined composition of modules that reflect the structure of ASTs.
\cite{sun_2020} propose a grammar dependent Transformer architecture for code generation and use convolutional layers to process natural language input and model hierarchical structure.
In contrast, we convey the compositional structure through the attention mechanism aided by a hierarchical positional encoding, not through parent feeding. 
Since we do not employ a grammar dependent architecture, with constrained decoding disabled, our approach can be applied to any other tree-structured data for which no abstract grammar is available.


For encoding nodes in a tree, we use the same notion of path as~\cite{shiv_2019}, and like theirs, our tree encoding for a path can be described as a composition of affine transformations.
However, with~\cite{shiv_2019}'s encoding, there is only one affine transformation per path segment encoding, which might limit the robustness of the encoding.
While their tree encoding of a node results from concatenating the one-hot encodings of the node's reversed path segments and weighting the encodings using a learned parameter, we employ the parameter-free sinusoidal encoding by~\cite{vaswani_2017} to encode each path segment.
Since the tree depth and width of~\cite{shiv_2019} encoding is fixed, it is not possible to apply it on arbitrarily wide trees, and sibling relationships can only be represented in binary trees.
These limitations do not apply to our encoding scheme, which can be applied to ordered trees of variable width.

\section{Data Representation}

\subsection{Input Data}
As input for training, we use several Python code corpora~\cite{ling_2016,conala_yin_2018,django_oda_2015} consisting of pairs of natural language instructions and corresponding code snippets.
We parse the Python code snippets into abstract syntax trees (ASTs) and generate a sequence of AST tokens in depth-first preorder, as depicted in Figure~\ref{fig:ast}.
\begin{figure*}[htbp]
\centering
\begin{subfigure}[b]{\columnwidth}
\centering
{
\input{ast}
}
\caption{AST}
\label{fig:ast}
\end{subfigure}
\begin{subfigure}[b]{\columnwidth}
\centering
\arrayrulecolor{black}
{
\input{edgepath}
}
\arrayrulecolor{black}
\caption{Edge paths}
\label{fig:epath}
\end{subfigure}
\caption{(a) Example of the abstract syntax tree (AST) of the code snippet~\texttt{a=10}. 
We create the AST sequence by adding the names of the object nodes in depth-first preorder:~\texttt{[sos, Module, Assign, Name, 'a', le, Num, 10, le, eos]}. (b) Object node edge paths for the AST of the snippet \texttt{a=10}. All edge paths have the same length. Each edge path consists of the path segments of its predecessor nodes.}
\label{fig:datarep}
\end{figure*}
The AST consists of object nodes (e.g.~\emph{Module}) and attribute nodes visualised by small squares and circles in the figure.
An attribute node is either a singleton (e.g.~\emph{value}) or an attribute list (e.g.~\emph{body}) that can point to several object nodes.
We integrated a special node for marking an attribute list end (\emph{le}) and special nodes for marking the start (\emph{sos}) and the end (\emph{eos}) of the code fragment.
An object node can be an inner node (e.g. \emph{Module}) or a leaf node (e.g. \emph{'a'}) and can have none, one or more attribute nodes.
Each attribute node has a defined order in its object node's list of attribute nodes.
An edge from an object node to an attribute indicates that the attribute belongs to that object node, and an edge from an attribute to an object node means that instances of that object node are allowed to appear as the value of the attribute.

We create the sequential representation of an AST by traversing the AST in depth-first preorder and adding the names of the object nodes to the AST sequence. 
An AST sequence consists of object node tokens and special nodes tokens.
If an object node is the last or the only node in an attribute list, we add a list end token after the object node's token in the AST sequence.
From the ASTs of the code samples corpora only used for training, we create a grammar model that captures the Python programming language syntax.
From the natural language input and the string literals extracted from the ASTs we create a common subword vocabulary using the BPE implementation of~\cite{sentencepiece_2018}.
From the tokens from the AST sequence, we generate a different word vocabulary without string literals.
Using a BPE segmentation for string literals has the advantage that we can effectively adjust the vocabulary size and reduce the number of tokens required to encode the input sequences~\cite{kudo_2018}.

\subsection{Grammar Model}
An ASDL grammar does not necessarily specify the exact set of valid ASTs, but only a context-free, parsing-optimized superset of ASTs.
As a result, the grammar is not sufficiently restrictive and therefore allows the generation of ASTs that do not correspond to any parsable program, or ASTs that are parsable but would not withstand simple static checks.
For instance, the Python ASDL allows spaces within variable identifiers or constant string or number literals on the left side of an assignment, which is an invalid syntax. CPython's parser (3.7) still allows constructs such as function calls or index operations with constants as targets to be parsed into ASTs, e.g. "1[2] = 3()", resulting in a static check error.

As an improvement, we propose to generate a restricted grammar model that allows only AST compositions (i.e.~attribute-child combinations) that occur in a corpus of Python code samples, e.g.~the train dataset of a task.
The construction of the grammar graph essentially induces a context-free generalization of the ASTs from the training dataset containing only parsable code samples.
This way, the likelihood of generating invalid ASTs can be reduced significantly.
The grammar graph defines a sub-language of the language defined by the ASDL by construction, so the knowledge contained in the ASDL is still preserved in the grammar graph and not discarded.
Indeed, the set of ASTs is restricted to what occurs in the training data, but this will more likely prevent the model from predicting invalid rather than valid AST compositions. The advantage of the restrictive grammar graph is that the model can only learn what it has seen in the data, ensuring the generation of parsable ASTs. 
In any case, the model does not allow the prediction of AST combinations that it has not seen, i.e., that did not occur in the training data. 

We define the grammar model as a bipartite labeled directed graph, whose nodes are the AST object nodes and attribute nodes.
The grammar graph, is used to compute edge paths for the tree encoding and can also be used to create constraint masks for limiting possible candidates from the current prediction during inference.

\section{Tree Encoding}
Since the Transformer model is invariant to the input order, due to its attention-based architecture, required position information needs to be explicitly included in the input sequence.
Thus,~\cite{vaswani_2017} encode positions in a sequence with a parameter-free scheme using sinusoidal functions of different frequencies and add these position encodings to the embeddings of the input sequence tokens of the first encoder and decoder layer.

However, flattening ASTs into an input sequence of AST tokens does not preserve the hierarchical order between the tree nodes.
In particular, nodes whose tokens are next to each other in the AST sequence can be far apart in the tree, namely if the preceding token in the AST sequence is the last node of a deep subtree. 
Conversely, the tokens of two neighboring nodes in the AST are far apart in the sequence when the first of two direct siblings is the root of a large subtree.

To counteract this problem, we propose a tree-based positional encoding built on~\cite{vaswani_2017}'s sinusoidal encoding scheme, which allows the attention mechanism to also attend over hierarchical positions in the input AST sequence.

\subsection{Edge Paths}
To define the tree encoding, we interpret the AST as an ordered tree \({T=(V_T, E_T, r_T)}\) consisting of a set of nodes \(V_T\) with \(n\) object nodes and \(m\) attribute nodes, a set of edges \(E_T\) and root \(r_T\).

A node \(v \in V_T\) is uniquely identified by an edge path \(p_{v}\) of a configurable length \(L\) if the depth of \(T\) is at most \(L\).
An edge path is not a path as in graph theory, i.e., a sequence of neighboring nodes, but rather structured like a path in a file system.
For each object and attribute node in the AST, the outgoing edges receive a consecutively numbered edge index \(idx\).
For example in Figure~\ref{fig:ast}, the first edge of the object node \emph{Assign} to the attribute node \emph{target} has index~1 and the second edge to the attribute node \emph{value} has index~2.

The edge path of a node consists of the edge indices of the edges on the path from the node to the root node. 
Edge paths shorter than \(L\) are padded with zeros and the root node's edge path consists only of zeros.
For example, the edge path of \emph{Assign} consists of the indices of the four edges on the path to the root node \emph{sos}.

Formally, each edge path \(p_{v} = (idx_{j,l})_{j=1}^{n}\vphantom{(idx_{j,l})}_{l=1}^{L}\)
consists of edge indices that lead from the target node upwards to the root node.
The \(l\)-th edge index of the edge path refers to the \(l\)-th edge of the (reverse) path from the target node up to the root.

\subsection{Encoding Scheme}
To encode the position of a node \(v\) in a tree \(T\), we apply the sinusoidal encoding by~\cite{vaswani_2017} to each edge index of the node's edge path \(p_{v}\) and compute the positional tree encoding \(TE_{j, l} = \left(EE_{idx_{j, l}, i\prime}\right)_{i\prime \in \{1, \dots, d_{idx}\}}\) with \(j \in \{1, \dots, n\}\) and  \(l \in \{1, \dots, L\}\) as follows:
\[ EE_{\mathit{idx_{j,l}},2i} = \mathit{sin}(\omega_{i} \cdot \mathit{idx_{j, l}})\] 
\vspace{-3.5ex}
\[ EE_{\mathit{idx_{j,l}},2i+1} = \mathit{cos}(\omega_{i} \cdot \mathit{idx_{j, l}})\]
\vspace{-3.5ex}
with \(\omega_{i} = 1 / 10000^{\frac{2i}{d_{idx}}} \) for \( i \in \{1, \dots, \frac{d_{idx}}{2}\} \). 
\vspace{3.5ex}

The parameter \(d_{idx}\) is the dimension of the positional encoding of an edge index \(idx_{j,l}\) and determines the number of encodings per edge index, i.e., \(\frac{d_{idx}}{2} \in \mathbb{N}\) sine and cosine pairs. This results in an encoding of length \(d_{idx} * L\) for an edge path, which also corresponds to \(d_m\), the size of the node's embedding.

The tree encoding \(TE_{v_j}\) of a node \(v_j\) is the concatenation of the encodings of its edge path indices:
\vspace{-1ex}
\[TE_{v_j} = TE_{j, 1} \| \dots \| TE_{j, L}\]
\vspace{-1ex}
with \(j\in\{1,\dots,n\}\).
\vspace{2ex}

\subsection{Encoding Properties}
The tree encoding entails the following properties, which result from the composition of the edge paths and the sine and cosine functions used to encode the edge paths: 

\paragraph{Uniqueness property}
The concatenation of the edge indices encodings of a edge path results in a unique tree encoding for a node \(v_j\).
This is because for a node \(v_j\), the combination of its edge path indices \(idx_{j,l}\) is unique and the sine (and cosine) function is injective on the products of its edge path indices and each frequency \(\omega_{i^\prime}\) due to the transcendality of~\(\pi\).

\paragraph{Shifting property}
The tree encoding of a child node \(v_j\) contains the first \((L-1)*d_{idx}\) dimensions of its parent node's tree encoding, shifted to the right by \(d_{idx}\) dimensions.
Two nodes are siblings if these shifted parent dimensions contain identical values for both nodes.
Similar to~\cite{shiv_2019}, the shifting property is intended to enable the attention mechanism to identify whether a node is an ancestor or a sibling of a node \(v_j\). 
However, relative distances between sibling nodes cannot be represented by shifting alone.

\paragraph{Linear combination property}
Due to the mathematical properties of the sine and cosine functions~\cite{vaswani_2017} hypothesizes that it is possible for the attention mechanism to attend to sequential order relationships, i.e., to learn relative positions.
We make use of this property to allow attending to sibling nodes by their relative positions.
From the positional encoding of an edge index \(idx_{j,l}\), the positional encoding of another edge index \(idx_{j,l}+k\) can be computed for any integer offset \(k\), using the sum rules for trigonometric functions.
The positional encoding of a siblings of a node \(v_j\) with an offset \(k < 0\) and l = 1 can be expressed as a linear combination of the sine and cosine values of the node \(v_j\) edge encoding. 
\section{Evaluation}

\subsection{Datasets}
For the evaluation we use the Python code generation benchmarks CoNaLa~\cite{conala_yin_2018} and Hearthstone (HS)~\cite{ling_2016} and for semantic parsing we use the datasets GEO~\cite{tang_mooney_2001} and Atis~\cite{dahl_1994}.

Hearthstone is a corpus for the automatic generation of code for cards in the trading card game HearthStone and consists of 665 Python classes, each representing one card.
CoNaLa is a Python corpus containing 2,379 curated and 593,891 mined NL-code pairs mined from the developer forum Stack Overflow. CoNaLa contains as NL intent real-world questions to diverse implementation topics instead of pseudo-code annotations.
GEO is a corpus of natural language questions about US geography and consists of 600 training and 280 test examples of NL-Prolog queries pairs.
Atis is a corpus of natural language queries for a flights database featuring 4473 training and 448 test examples of NL-lamba-calculus pairs.

\subsection{Metrics}
As evaluation metrics, we use the BLEU score as implemented by~\cite{conala_yin_2018}, the exact match accuracy and the token- and sequence-level prefix precision and recall.

The BLEU score measures the similarity between the generated code and the reference code in terms of n-grams.
We compute the token-level BLEU score on normalized and tokenized code by determining the precision on n-grams, then take the geometric mean of the precisions, and apply a brevity penalty for predictions shorter than the expected code.

Exact match accuracy measures the ratio of predictions that were predicted without a single error.
This ensures the semantic correctness of the code, but it is very conservative in that it doesn‘t allow the slightest syntactic variation, and that no distinction is made between a prediction that is 0\% correct and one that is 99\% correct.

Token- and sequence-level prefix precision and recall interpolate exact match accuracies based on the idea that if some prefix was predicted correctly, the model didn‘t make a mistake up to that point, so it is alright to give the model some credit for that.
On token level, it measures the ratio of tokens in correct prefixes, while on sequence level, it measures the average percentage of the longest correct prefix, ignoring snippet length.
The prefix-based metrics also partially measure semantical correctness because the expected snippet is by definition semantically correct.

\subsection{Experimental Setup}
We performed the experiments on a cluster of IBM AC922 nodes with dual Power9 CPUs (2.80-3.10 GHz, 22 cores each), 256 GB RAM and 6 Nvidia Volta V100 accelerators with 32 GB RAM. And on nodes with AMD EPYC CPUs (2.3 GHz, 24 cores each), 1 TB RAM and 8 A100 GPUs with 40 GB RAM per node.

In the following experiments, unless specified otherwise, we train each task for 500 epochs with batch size~\(15\), learning rate~\(0.0001\),~\(6\) encoder and~\(6\) to~\(8\) decoder layers,~\(16\) attention heads, model dimension~\(512\), maximum tree height~\(32\) and feed-forward dimension~\(2048\).
For inference, we set the number of beams to~\(k=30\) and use a maximum beam length of 250 tokens.

\subsection{Evaluation Results}

We performed the evaluation on two tasks, namely semantic parsing on GEO and ATIS and code generation on CoNaLa and Hearthstone.
In both cases, the goal is to generate formal meaning representations from natural language input, that is, lambda-calculus expressions or Python code.

\begin{table*}[htbp]
\centering
{
\input{bleu_sota}
}
\caption{Comparison BLEU score with the state of the art on CoNaLa and Hearthstone.}
\label{tab:bleusota}
\end{table*}
\begin{table*}[htbp]
\centering
{
\input{em_sota}
}
\caption{Comparison EM accuracy with the state of the art on GEO and Atis.}
\label{tab:emota}
\end{table*}
\begin{table*}[htbp]
\centering
{
\input{em_lcp}
}
\caption{Exact match accuracy and longest common prefix on Hearthstone. }
\label{tab:emlcp}
\end{table*}

We scored about 18\% and 70\% BLEU score for the CoNaLa and Hearthstone benchmarks, respectively (cf. table~\ref{tab:bleusota}).
State of the art approaches have more sophisticated implementations that include dynamic composite neural architecture, additional embedded information, pre-training, or re-ranking.

Our results on the semantic parsing datasets GEO and ATIS are listed in table~\ref{tab:emota}.
On ATIS we achieved an exact match accuracy of 86,1\%, which is comparable to all our baselines except for the leading approach.
The exact match accuracy on GEO is below the baselines, which we conjecture to be caused by the small amount of training data.

To test whether the tree-encoded Transformer learns to predict the AST structure correctly, we looked at the exact match accuracy and token- and sequence-level precision and recall, as shown in table~\ref{tab:emlcp}.

We masked all string literals and computed the longest common prefix between the best predicted and the expected AST sequence for each sample from the test dataset.
We do this on a model we trained with tree encoding and then on a model we trained with sequential encoding.

With the exclusion of string literals, the prefix precision and recall jump from about 25\% to about 50\% with both sequential and tree encoding. 
From the prefix analysis, we can take away that the string literals have a significant impact on the quality of the prediction and that longer sequences are more difficult to predict. 
What we also found is that tree encoding gives an improvement of up to 3.0\% when excluding string literals over sequential encoding. 

\section{Conclusion}
\label{sec:conclusion}
We propose a Transformer-based architecture with a tree-based positional encoding and constrained decoding based on a grammar model derived from training data.
We evaluate it on four different datasets.
While we do not surpass state-of-the-art methods, we see relative improvement of applying tree encoding over sequential encoding.

In the future, we will work on improved versions of training with constraint masks and grammar models that respect larger contexts than only the immediate containing attribute.
Also, we will further investigate attention-based mechanisms for generating tree structures, also in the context of domain-specific languages as generation target.

\section{Acknowledgments}
This work was supported by the German Federal Ministry of Education and Research (BMBF, 01IS18026A-D) by funding the competence center for Big Data and AI "ScaDS.AI Dresden/Leipzig". The authors gratefully acknowledge the GWK support for funding this project by providing computing time through the Center for Information Services and HPC (ZIH) at TU Dresden.

\bibliography{literature}

\end{document}

%% file: datasets.tex
\begin{tabular}{lll}
\textbf{Dataset}    & \textbf{NL Specification}                                                                                                                                                                                                                                                                                                                                   & \textbf{Formal Meaning Representation}                                                                                                                                                                                                                                                                                                                                             \\ 
\hline\hline
\vcell{Atis}        & \vcell{flight from ci0 mn0 dn0}                                                                                                                                                                                                                                                                                                                             & \vcell{\begin{tabular}[b]{@{}l@{}}\textit{( lambda \$0 e ( and ( flight \$0 )}\textit{~ ( from \$0 ci0 )}\\\textit{~~~( day\_number \$0 dn0 )}\textit{ ( month \$0 mn0 ) )}\end{tabular}}                                                                                                                                                                                \\[-\rowheight]
\printcelltop       & \printcelltop                                                                                                                                                                                                                                                                                                                                               & \printcelltop                                                                                                                                                                                                                                                                                                                                                                      \\ 
\hline
\vcell{Geo}         & \vcell{\begin{tabular}[b]{@{}l@{}}What is the area of the state with \\the smallest population density?\end{tabular}}                                                                                                                                                                                                                                       & \vcell{\begin{tabular}[b]{@{}l@{}}\textit{( area:i ( argmin \textbackslash{}\$0}\\\textit{~~~ ( state:t \textbackslash{}\$0 ) ( density:i \textbackslash{}\$0 ) ) )}\end{tabular}}                                                                                                                                                                                                 \\[-\rowheight]
\printcelltop       & \printcelltop                                                                                                                                                                                                                                                                                                                                               & \printcelltop                                                                                                                                                                                                                                                                                                                                                                      \\ 
\hline
\vcell{CoNaLa}      & \vcell{How to load a csv file?}                                                                                                                                                                                                                                                                                                                             & \vcell{\begin{tabular}[b]{@{}l@{}}\textit{import csv}\\\textit{with open('file.csv') as csvfile:}\\\textit{~ reader = csv.DictReader(csvfile)}\\\end{tabular}}  \\[-\rowheight]
\printcelltop       & \printcelltop                                                                                                                                                                                                                                                                                                                                               & \printcelltop                                                                                                                                                                                                                                                                                                                                                                      \\ 
\hline
\vcell{Hearthstone} & \vcell{\begin{tabular}[b]{@{}l@{}}Treant NAME\_END 2 ...~ \\Minion TYPE\_END\\Druid~ PLAYER\_CLS\_END\\NIL RACE\_END NIL \\ RARITY\_END NIL\end{tabular}} & \vcell{\begin{tabular}[b]{@{}l@{}}\textit{class Treant(MinionCard):}\\\textit{~def \_\_init\_\_(self):}\\\textit{~ super().\_\_init\_\_("Treant", 1,}\\\textit{~~ CHARACTER\_CLASS.DRUID,}\\\textit{~~ CARD\_RARITY.COMMON)}\\\textit{~def create\_minion(self, \_):}\\\textit{~ return Minion(2, 2)}\end{tabular}}                                                                \\[-\rowheight]
\printcelltop       & \printcelltop                                                                                                                                                                                                                                                                                                                                               & \printcelltop                                                                                                                                                                                                                                                                                                                                                                      \\
\hline
\end{tabular}

%% file: ast.tex
\tikzset{every picture/.style={line width=0.75pt}} 

\begin{tikzpicture}[x=0.75pt,y=0.75pt,yscale=-1,xscale=1]

\draw   (70,54.5) -- (120.5,54.5) -- (120.5,68) -- (70,68) -- cycle ;
\draw  [fill={rgb, 255:red, 0; green, 0; blue, 0 }  ,fill opacity=1 ] (90.67,78.39) -- (100.28,78.39) -- (100.28,88) -- (90.67,88) -- cycle ;
\draw    (95.24,68.55) -- (95.3,78.04) ;
\draw    (94.57,21.89) -- (94.64,31.37) ;
\draw   (90.14,35.87) .. controls (90.14,33.39) and (92.15,31.37) .. (94.64,31.37) .. controls (97.12,31.37) and (99.14,33.39) .. (99.14,35.87) .. controls (99.14,38.36) and (97.12,40.37) .. (94.64,40.37) .. controls (92.15,40.37) and (90.14,38.36) .. (90.14,35.87) -- cycle ;
\draw    (94.64,40.37) -- (94.37,53.58) ;
\draw [shift={(94.33,55.58)}, rotate = 271.14] [color={rgb, 255:red, 0; green, 0; blue, 0 }  ][line width=0.75]    (10.93,-3.29) .. controls (6.95,-1.4) and (3.31,-0.3) .. (0,0) .. controls (3.31,0.3) and (6.95,1.4) .. (10.93,3.29)   ;
\draw   (70.14,103.25) -- (120.64,103.25) -- (120.64,116.75) -- (70.14,116.75) -- cycle ;
\draw    (95.8,88.04) -- (95.54,101.25) ;
\draw [shift={(95.5,103.25)}, rotate = 271.14] [color={rgb, 255:red, 0; green, 0; blue, 0 }  ][line width=0.75]    (10.93,-3.29) .. controls (6.95,-1.4) and (3.31,-0.3) .. (0,0) .. controls (3.31,0.3) and (6.95,1.4) .. (10.93,3.29)   ;
\draw   (106.2,12.2) -- (160.5,12.2) -- (160.5,31.75) ;
\draw   (50.96,123.46) -- (51.04,110.18) -- (70.18,110.29) ;
\draw  [fill={rgb, 255:red, 0; green, 0; blue, 0 }  ,fill opacity=1 ] (45.95,123.53) -- (55.56,123.53) -- (55.56,133.14) -- (45.95,133.14) -- cycle ;
\draw    (160.5,40.75) -- (160.38,54.39) ;
\draw [shift={(160.36,56.39)}, rotate = 270.52] [color={rgb, 255:red, 0; green, 0; blue, 0 }  ][line width=0.75]    (10.93,-3.29) .. controls (6.95,-1.4) and (3.31,-0.3) .. (0,0) .. controls (3.31,0.3) and (6.95,1.4) .. (10.93,3.29)   ;
\draw    (50.87,131.33) -- (50.6,148.54) ;
\draw [shift={(50.57,150.54)}, rotate = 270.9] [color={rgb, 255:red, 0; green, 0; blue, 0 }  ][line width=0.75]    (10.93,-3.29) .. controls (6.95,-1.4) and (3.31,-0.3) .. (0,0) .. controls (3.31,0.3) and (6.95,1.4) .. (10.93,3.29)   ;
\draw   (25.57,150.36) -- (76.07,150.36) -- (76.07,163.86) -- (25.57,163.86) -- cycle ;
\draw    (50.81,164.41) -- (50.87,173.9) ;
\draw   (46.37,178.4) .. controls (46.37,175.91) and (48.39,173.9) .. (50.87,173.9) .. controls (53.36,173.9) and (55.37,175.91) .. (55.37,178.4) .. controls (55.37,180.88) and (53.36,182.9) .. (50.87,182.9) .. controls (48.39,182.9) and (46.37,180.88) .. (46.37,178.4) -- cycle ;
\draw    (50.87,182.9) -- (50.61,196.11) ;
\draw [shift={(50.57,198.11)}, rotate = 271.14] [color={rgb, 255:red, 0; green, 0; blue, 0 }  ][line width=0.75]    (10.93,-3.29) .. controls (6.95,-1.4) and (3.31,-0.3) .. (0,0) .. controls (3.31,0.3) and (6.95,1.4) .. (10.93,3.29)   ;
\draw    (138.95,163.13) -- (139.02,172.61) ;
\draw   (134.52,177.11) .. controls (134.52,174.63) and (136.53,172.61) .. (139.02,172.61) .. controls (141.5,172.61) and (143.52,174.63) .. (143.52,177.11) .. controls (143.52,179.6) and (141.5,181.61) .. (139.02,181.61) .. controls (136.53,181.61) and (134.52,179.6) .. (134.52,177.11) -- cycle ;
\draw    (139.02,181.61) -- (138.75,194.82) ;
\draw [shift={(138.71,196.82)}, rotate = 271.14] [color={rgb, 255:red, 0; green, 0; blue, 0 }  ][line width=0.75]    (10.93,-3.29) .. controls (6.95,-1.4) and (3.31,-0.3) .. (0,0) .. controls (3.31,0.3) and (6.95,1.4) .. (10.93,3.29)   ;
\draw   (156,36.25) .. controls (156,33.76) and (158.01,31.75) .. (160.5,31.75) .. controls (162.99,31.75) and (165,33.76) .. (165,36.25) .. controls (165,38.74) and (162.99,40.75) .. (160.5,40.75) .. controls (158.01,40.75) and (156,38.74) .. (156,36.25) -- cycle ;
\draw   (120.64,109.25) -- (139.6,109.25) -- (139.6,126.75) ;
\draw    (139.5,135.75) -- (139.64,148.83) ;
\draw [shift={(139.67,150.83)}, rotate = 269.37] [color={rgb, 255:red, 0; green, 0; blue, 0 }  ][line width=0.75]    (10.93,-3.29) .. controls (6.95,-1.4) and (3.31,-0.3) .. (0,0) .. controls (3.31,0.3) and (6.95,1.4) .. (10.93,3.29)   ;
\draw   (135,131.25) .. controls (135,128.76) and (137.01,126.75) .. (139.5,126.75) .. controls (141.99,126.75) and (144,128.76) .. (144,131.25) .. controls (144,133.74) and (141.99,135.75) .. (139.5,135.75) .. controls (137.01,135.75) and (135,133.74) .. (135,131.25) -- cycle ;
\draw  [color={rgb, 255:red, 0; green, 0; blue, 0 }  ,draw opacity=1 ] (114.33,151.08) -- (164.83,151.08) -- (164.83,164.58) -- (114.33,164.58) -- cycle ;
\draw   (54.31,128.25) -- (94.17,128.25) -- (94.17,145.75) ;
\draw    (94.17,128.25) -- (94.04,148.39) ;
\draw [shift={(94.02,150.39)}, rotate = 270.37] [color={rgb, 255:red, 0; green, 0; blue, 0 }  ][line width=0.75]    (10.93,-3.29) .. controls (6.95,-1.4) and (3.31,-0.3) .. (0,0) .. controls (3.31,0.3) and (6.95,1.4) .. (10.93,3.29)   ;
\draw  [fill={rgb, 255:red, 215; green, 215; blue, 215 }  ,fill opacity=1 ] (94.5,1.6) -- (107.4,12) -- (94.5,22.4) -- (81.6,12) -- cycle ;
\draw   (100.31,83.25) -- (160.6,83.25) -- (160.6,83.83) ;
\draw    (161.17,83.25) -- (161.04,103.39) ;
\draw [shift={(161.02,105.39)}, rotate = 270.37] [color={rgb, 255:red, 0; green, 0; blue, 0 }  ][line width=0.75]    (10.93,-3.29) .. controls (6.95,-1.4) and (3.31,-0.3) .. (0,0) .. controls (3.31,0.3) and (6.95,1.4) .. (10.93,3.29)   ;

\draw (94.94,60.57) node   [align=left] {{\scriptsize Module}};
\draw (94.5,13) node   [align=left] {{\scriptsize sos}};
\draw (94.4,110.29) node   [align=left] {{\scriptsize Assign}};
\draw (50.58,156.86) node   [align=left] {{\scriptsize Name}};
\draw (50.58,208.24) node   [align=left] {{\scriptsize 'a'}};
\draw (139.44,207.79) node   [align=left] {{\scriptsize 10}};
\draw (35.37,178.4) node [anchor=west] [inner sep=0.75pt]   [align=left] {{\tiny id}};
\draw (124.08,177.17) node [anchor=west] [inner sep=0.75pt]   [align=left] {{\tiny n}};
\draw (21.88,128.37) node [anchor=west] [inner sep=0.75pt]   [align=left] {{\tiny target}};
\draw (70.21,36.84) node [anchor=west] [inner sep=0.75pt]   [align=left] {{\tiny start}};
\draw (71.28,84.87) node [anchor=west] [inner sep=0.75pt]   [align=left] {{\tiny body}};
\draw (140.01,36.7) node [anchor=west] [inner sep=0.75pt]   [align=left] {{\tiny end}};
\draw (160.92,67.17) node   [align=left] {{\scriptsize eos}};
\draw (112.61,131.5) node [anchor=west] [inner sep=0.75pt]   [align=left] {{\tiny value}};
\draw (138.58,157.83) node   [align=left] {{\scriptsize Num}};
\draw (93.58,160.83) node   [align=left] {{\scriptsize le}};
\draw (161.25,115.83) node   [align=left] {{\scriptsize le}};
\draw (144,170) node [anchor=west] [inner sep=0.75pt]   [align=left] {{\tiny 1}};
\draw (144,187) node [anchor=west] [inner sep=0.75pt]   [align=left] {{\tiny 1}};
\draw (144,138) node [anchor=west] [inner sep=0.75pt]   [align=left] {{\tiny 1}};
\draw (144,117) node [anchor=west] [inner sep=0.75pt]   [align=left] {{\tiny 2}};
\draw (165,93) node [anchor=west] [inner sep=0.75pt]   [align=left] {{\tiny 2}};
\draw (99,93) node [anchor=west] [inner sep=0.75pt]   [align=left] {{\tiny 1}};
\draw (99,72.17) node [anchor=west] [inner sep=0.75pt]   [align=left] {{\tiny 1}};
\draw (99,26) node [anchor=west] [inner sep=0.75pt]   [align=left] {{\tiny 1}};
\draw (99,46) node [anchor=west] [inner sep=0.75pt]   [align=left] {{\tiny 1}};
\draw (165,26) node [anchor=west] [inner sep=0.75pt]   [align=left] {{\tiny 2}};
\draw (165,46) node [anchor=west] [inner sep=0.75pt]   [align=left] {{\tiny 1}};
\draw (55,170) node [anchor=west] [inner sep=0.75pt]   [align=left] {{\tiny 1}};
\draw (55,187) node [anchor=west] [inner sep=0.75pt]   [align=left] {{\tiny 1}};
\draw (55,138) node [anchor=west] [inner sep=0.75pt]   [align=left] {{\tiny 1}};
\draw (99,138) node [anchor=west] [inner sep=0.75pt]   [align=left] {{\tiny 2}};
\draw (55,117) node [anchor=west] [inner sep=0.75pt]   [align=left] {{\tiny 1}};

\end{tikzpicture}

%% file: edgepath.tex
\begin{tabular}{ll}
\textbf{Node} & \textbf{Path}                \\ 
\hline
sos           & {[}0, 0, 0, 0, 0, 0, 0, 0, 0, 0]  \\
Module        & {[}1, 1, 0, 0, 0, 0, 0, 0, 0, 0]  \\
Assign        & {[}1, 1, 1, 1, 0, 0, 0, 0, 0, 0]  \\
Name          & {[}1, 1, 1, 1, 1, 1, 0, 0, 0, 0]  \\
'a'           & {[}1, 1, 1, 1, 1, 1, 1, 1, 0, 0]  \\
le            & {[}2, 1, 1, 1, 1, 1, 0, 0, 0, 0]  \\
Num           & {[}1, 2, 1, 1, 1, 1, 0, 0, 0, 0]  \\
10            & {[}1, 1, 1, 2, 1, 1, 1, 1, 0, 0]  \\
le            & {[}2, 1, 1, 1, 0, 0, 0, 0, 0, 0]  \\
eos           & {[}1, 2, 0, 0, 0, 0, 0, 0, 0, 0] 
\end{tabular}

%% file: bleu_sota.tex
\begin{tabular}{p{25mm}p{18mm}p{10mm}p{5mm}p{30mm}p{18mm}p{10mm}}

{\cellcolor[HTML]{EFEFEF}Authors}                                                                           & {\cellcolor[HTML]{EFEFEF}Name}                                                          & {\cellcolor[HTML]{EFEFEF}BLEU} &  & {\cellcolor[HTML]{EFEFEF}Authors}                                                           & {\cellcolor[HTML]{EFEFEF}Name}      & {\cellcolor[HTML]{EFEFEF}BLEU} \\  \cline{1-3} \cline{5-7} 

\begin{tabular}[c]{@{}l@{}}Yin et al. 2019\end{tabular}                        & tranX                                                         & 24,4\%                       &  & \begin{tabular}[c]{@{}l@{}}Yin et al.  2019\end{tabular}        & tranX     & 75,8\%                       \\ \cline{1-3} \cline{5-7} 
\begin{tabular}[c]{@{}l@{}}Yin et al. 2019\end{tabular}                        & \begin{tabular}[c]{@{}l@{}}tranX\\ + rerank\end{tabular}      & 30,1\%                       &  & \begin{tabular}[c]{@{}l@{}}Hayati et al. 2018\end{tabular}     & ReCode    & 78,4\%                       \\  \cline{1-3} \cline{5-7} 
\begin{tabular}[c]{@{}l@{}}Xu et al. 2020\end{tabular} & \begin{tabular}[c]{@{}l@{}}tranX\\ + pre-trained\end{tabular} & \textbf{32,3}\%                       &  & \begin{tabular}[c]{@{}l@{}}Rabinovich et al. 2017\end{tabular} & ASNs      & 79,2\%                       \\ \cline{1-3} \cline{5-7} 
\multicolumn{3}{l}{}                                                                                                                                                           &  & \begin{tabular}[c]{@{}l@{}}Sun et al. 2020\end{tabular}        & TreeGen-B & \textbf{81,8}\%                       \\ \cline{1-3} \cline{5-7} 
\multicolumn{2}{l}{Our system}                                                                                          & 18,1\%                       &  & \multicolumn{2}{l}{Our system}                       & 70,7\%                       \\ \cline{1-3} \cline{5-7}

\multicolumn{3}{c}{(a) CoNaLa}                                                                                                                            &  & \multicolumn{3}{c}{(b) Hearthstone}                                                 
\end{tabular}

%% file: em_sota.tex
\renewcommand{\arraystretch}{1.5}
\begin{tabular}{p{35mm}p{35mm}p{20mm}p{20mm}}
\cellcolor[HTML]{EFEFEF}Authors                         & \cellcolor[HTML]{EFEFEF}Name                   & \cellcolor[HTML]{EFEFEF}GEO & \cellcolor[HTML]{EFEFEF}Atis \\ \hline
Rabinovich et al. 2017          & ASNs          & 87,1\%                      & 85,9\%                       \\ \hline
Yin \& Neubig 2018              & tranX                  & 88,2\%                      & 86,2\%                       \\ \hline
Dong \& Lapata 2018             & oracle sketch          & \textbf{93,9}\%                      & \textbf{95,1}\%                       \\ \hline
Shiv et al. 2019                & Seq2Tree Tform         & 84,6\%                      & 86,4\%                       \\ \hline
\multicolumn{2}{l}{Our system} &  80,4\%                      &  86,1\%                       \\ \hline
\end{tabular}                             

%% file: em_lcp.tex
\begin{tabular}{p{9mm}p{9mm}p{9mm}p{9mm}p{9mm}p{9mm}p{9mm}p{9mm}p{9mm}p{9mm}p{9mm}p{9mm}}
\cellcolor[HTML]{EFEFEF}\begin{tabular}[c]{@{}l@{}}Pos. \\ Enc.\end{tabular} & \begin{tabular}[c]{@{}l@{}}Incl.\\ Str.\\ Lit.\end{tabular} & \cellcolor[HTML]{EFEFEF}\begin{tabular}[c]{@{}l@{}}EM\\ Acc.\end{tabular} & \begin{tabular}[c]{@{}l@{}}Seq\\ vs.\\ Tree\end{tabular}     & \cellcolor[HTML]{EFEFEF}\begin{tabular}[c]{@{}l@{}}Seq.\\ Recall\end{tabular} & \begin{tabular}[c]{@{}l@{}}Seq\\ vs.\\ Tree\end{tabular}     & \cellcolor[HTML]{EFEFEF}\begin{tabular}[c]{@{}l@{}}Token\\ Recall\end{tabular} & \begin{tabular}[c]{@{}l@{}}Seq\\ vs.\\ Tree\end{tabular}     & \cellcolor[HTML]{EFEFEF}\begin{tabular}[c]{@{}l@{}}Seq.\\ Prec.\end{tabular} & \begin{tabular}[c]{@{}l@{}}Seq\\ vs.\\ Tree\end{tabular}     & \cellcolor[HTML]{EFEFEF}\begin{tabular}[c]{@{}l@{}}Token\\ Prec.\end{tabular} & \begin{tabular}[c]{@{}l@{}}Seq\\ vs.\\ Tree\end{tabular} \\ 
\hline
\hline
\multicolumn{1}{c}{Seq}                                                    & \multicolumn{1}{l}{yes}                                    & \multicolumn{1}{c}{4,5\%}                                                & \multicolumn{1}{c}{}                                        & \multicolumn{1}{c}{26,9\%}                                                   & \multicolumn{1}{c}{}                                        & \multicolumn{1}{c}{23,0\%}                                                    & \multicolumn{1}{c}{}                                        & \multicolumn{1}{c}{28,4\%}                                                  & \multicolumn{1}{c}{}                                        & \multicolumn{1}{c}{26,4\%}                                                   &                                                          \\ 
\multicolumn{1}{c}{Seq}                                                    & \multicolumn{1}{l}{no}                                     & \multicolumn{1}{c}{10,6\%}                                               & \multicolumn{1}{c}{\multirow{-2}{*}{}}                      & \multicolumn{1}{c}{48,9\%}                                                   & \multicolumn{1}{c}{\multirow{-2}{*}{}}                      & \multicolumn{1}{c}{42,5\%}                                                    & \multicolumn{1}{c}{\multirow{-2}{*}{}}                      & \multicolumn{1}{c}{52,2\%}                                                  & \multicolumn{1}{c}{\multirow{-2}{*}{}}                      & \multicolumn{1}{c}{48,7\%}                                                   & \multirow{-2}{*}{}                                       \\ \hline
\multicolumn{1}{c}{Tree}                           & \multicolumn{1}{l}{yes}                                    & \multicolumn{1}{c}{7,6\%}                                                & \multicolumn{1}{c}{\cellcolor[HTML]{FFFFFF}\textbf{+3,0\%}} & \multicolumn{1}{c}{28,9\%}                                                   & \multicolumn{1}{c}{\cellcolor[HTML]{FFFFFF}\textbf{+2,0\%}} & \multicolumn{1}{c}{23,7\%}                                                    & \multicolumn{1}{c}{\cellcolor[HTML]{FFFFFF}\textbf{+0,7\%}} & \multicolumn{1}{c}{30,5\%}                                                  & \multicolumn{1}{c}{\cellcolor[HTML]{FFFFFF}\textbf{+2,1\%}} & \multicolumn{1}{c}{27,9\%}                                                   & \cellcolor[HTML]{FFFFFF}\textbf{+1,4\%}                  \\ 
\multicolumn{1}{c}{Tree}                           & \multicolumn{1}{l}{no}                                     & \multicolumn{1}{c}{12,1\%}                                               & \multicolumn{1}{c}{\cellcolor[HTML]{FFFFFF}\textbf{+1,5\%}} & \multicolumn{1}{c}{50,0\%}                                                   & \multicolumn{1}{c}{\cellcolor[HTML]{FFFFFF}\textbf{+1,1\%}} & \multicolumn{1}{c}{43,7\%}                                                    & \multicolumn{1}{c}{\cellcolor[HTML]{FFFFFF}\textbf{+1,2\%}} & \multicolumn{1}{c}{53,9\%}                                                  & \multicolumn{1}{c}{\cellcolor[HTML]{FFFFFF}\textbf{+1,7\%}} & \multicolumn{1}{c}{51,3\%}                                                   & \cellcolor[HTML]{FFFFFF}\textbf{+2,7\%}                  \\ 
\end{tabular}